\documentclass[letterpaper, 10 pt, conference]{ieeeconf}  

\IEEEoverridecommandlockouts                              

\usepackage{graphics} 
\usepackage{epsfig} 
\usepackage{mathptmx} 
\usepackage{times} 
\usepackage{amsmath} 
\usepackage{amssymb}  
\usepackage{xcolor}
\usepackage{colortbl}
\usepackage{amsmath}
\newcommand\norm[1]{\left\lVert#1\right\rVert}

\usepackage[top=1in, left=0.75in, bottom=0.75in, right=0.75in]{geometry}

\title{\LARGE \bf
FusionMapping: Learning Depth Prediction with \\ Monocular Images and 2D Laser Scans
} 

\author{Peng Yin$^{1}$, Jianing Qian$^{1}$, Yibo Cao$^{1}$, David Held$^{1}$ and Howie Choset$^{1}$
\thanks{* The first two authors contributed equally to this work.}
\thanks{$^{1}$ P. Yin, J. Qian, Y. Cao, D. Held, H. Choset are with the Robotics Institute at Carnegie Mellon University, Pittsburgh, PA 15213, USA  ({\tt\small pyin2,jianingq,yiboc,dheld,choset})@andrew.cmu.edu}%
}

\begin{document}

\maketitle
\thispagestyle{empty}
\pagestyle{empty}

\begin{abstract}
    Acquiring accurate three-dimensional depth information conventionally requires expensive multibeam LiDAR devices.
    Recently, researchers have developed a less expensive option by predicting depth information from two-dimensional color imagery.
    However, there still exists a substantial gap in accuracy between depth information estimated from two-dimensional images and real LiDAR point-cloud.
    In this paper, we introduce a fusion-based depth prediction method, called FusionMapping. This is the first method that fuses colored imagery and two-dimensional laser scan to estimate depth information.
    More specifically, we propose an autoencoder-based depth prediction network and a novel point-cloud refinement network for depth estimation.
    We analyze the performance of our FusionMapping approach on the KITTI LiDAR odometry dataset and an indoor mobile robot system.
    The results show that our introduced approach estimates depth with better accuracy when compared to existing methods.
 
\end{abstract}

\section{INTRODUCTION}
\begin{figure*}[t]
    	\centering
    	\includegraphics[width=1.0\linewidth]{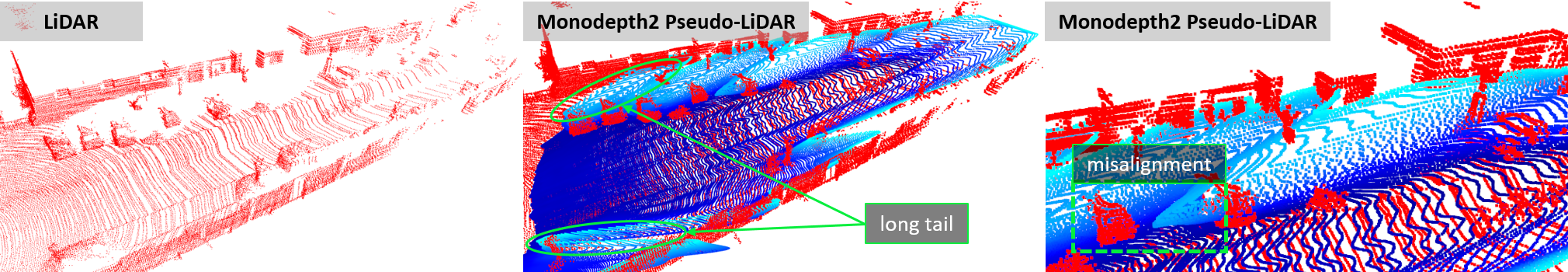}
    	\caption{3D point-cloud refinement. 
    	The left figure shows the original LiDAR point-cloud;
    	the middle figure shows that the depth information estimated from monocular~\cite{PSEUDO:pseudo_detection} exist \textit{long tails} and \textit{misalignment} problem, the red points are the ground truth and blue points are the estimated point-cloud;
    	the last figure shows that our proposed approach could deal with the above problem challenges with the assistance of 2D laser scans.}
    	\label{fig:problem}
    \end{figure*}

    \noindent Three-dimensional perception is an important capability for safe navigation, such as in autonomous driving~\cite{Auto:urmson2008autonomous,Auto:larson2019autonomous} and with mobile robots~\cite{Mobile:jinno20193d}.
    Several tasks, including three-dimensional object detection~\cite{3DPerc:liang2019multi}, tracking~\cite{3DPerc:zarzar2019efficient}, and scene segmentation~\cite{3DSegL:wu2018squeezesegv2} rely heavily on accurate and expensive multibeam LiDAR sensors.
    Unfortunately, the high cost of LiDAR sensors prohibits their mass utilization on mobile systems. 
    Recently, learning-based three-dimensional point-cloud prediction approaches~\cite{PSEUDO:pseudo_lidar,PSEUDO:pseudo_lidar++,PSEUDO:pseudo_detection} show potential in reducing the dependency on LiDAR devices by predicting depth information from 2D images.
    However, learning-based point-cloud predictions from images still fall short of those generated by high resolution LiDAR devices. 
    Figure~\ref{fig:problem} presents some examples of issues encountered by current approaches of image-based depth prediction:  
    1) the predicted point-cloud often extends beyond object boundaries, namely long tails (Figure~\ref{fig:problem} (middle));
    2) there is often a local misalignment with respect to the LiDAR point-cloud (Figure~\ref{fig:problem} (right)).

    In comparison to 3D LiDAR device, 2D (planar) LiDAR devices, which use a single beam instead of multiple beams, are much more cost-efficient to Automated Guided Vehicles (AGV) and service robot systems.
    To address the above problems in learning-based depth prediction, we introduce FusionMapping, a new depth prediction approach to reduce the discrepancy between high resolution LiDAR point-clouds and estimated point-clouds.
    This approach integrates monocular images and 2D laser scans to estimate accurate depth information.
    Specifically, we use a Depth Estimation Network ($F_{est}$) with image-laser fusion layers, namely early fusion layers, to generate depth prediction.
    
    In addition, we propose a Depth Refinement Network ($F_{refine}$) to refine the estimated depth prediction with 2D laser scans.
    In the early fusion layers of the $F_{est}$ network, directly fusing a 2D laser scan information onto a front-view monocular image is difficult due to their different perspectives.
    We handle this difficulty by transforming the 2D laser scan into a 3D-Mask (a feature fusion operation as shown in Figure~\ref{fig:early_fusion}, explanation of this operation is in Section. III (b)), then feeding this 3D-Mask, along with the original image, into $F_{est}$.
    With this preliminary depth prediction, the depth refinement network ($F_{refine}$) further refines the generated depth prediction.
    
    To summarize, we make the following contributions:
    \begin{itemize}
        \item An image-laser fusion-based depth prediction network, where 2D laser information is provided as environmental geometry constraints to refine the depth estimation from images.
        \item A Depth Estimation Network and a novel Depth Refinement Network, which are integrated with image-laser fusion and 2D laser refinement modules.
        \item Extensive quantitative and qualitative evaluation on the \textit{KITTI odometry} dataset and on a mobile robot system. The experiment results show that the proposed method significantly outperforms single monocular-image based prediction methods.
    \end{itemize}
    

\section{Related Work}
    Depth prediction has been well-studied in the past few years.
    In the following sub-sections, we will briefly investigate image-based and fusion-based depth prediction approaches.
    
    \subsection{Image-based Depth Prediction}
        Conventional depth prediction approaches mainly use hand-crafted features~\cite{SLAM:newcombe2011kinectfusion,SLAM:salas2014dense}. 
        For instance, Horncek et al. \cite{Tradition:DepthSuperRes} apply patches of 3D points to obtain depth information at a target resolution. Karsch et al. \cite{Tradition:DepthTransfer} use a non-parametric approach by adding optical flow information into the depth prediction pipeline.
        Nevertheless, these conventional approaches are not stable, since hand-crafted features are easily affected by illumination or texture changes.
    
        Recently, learning-based approaches have outperformed conventional methods on both stereo and monocular depth estimation. For example, many end-to-end learning approaches can directly conduct depth prediction on images \cite{Supervised:single-image}. 
        Both Liu et al. \cite{Supervised:cnn} and Fu et al. \cite{Supervised:DORN} perform supervised depth estimation with a single RGB image under specific network adjustments. 
        Godard et al. \cite{PSEUDO:monodepth} use unsupervised learning for monocular depth prediction by constructing a network to perform image reconstruction from stereo image pairs.
        Zhou et al. \cite{PSEUDO:ego-motion} propose a  unsupervised learning framework for depth prediction with video frames as input.
        Godard et al. \cite{PSEUDO:monodepth2} improve the depth prediction performance for both stereo pairs and monocular images using a photometric reconstruction loss and an auto-masking approach.
        Recent works also use depth estimation to generate pseudo-LiDAR; Wang et al. \cite{PSEUDO:pseudo_lidar} and Weng et al. \cite{PSEUDO:pseudo_detection} integrate a camera matrix with depth prediction to generate pseudo-LiDAR point-clouds for 3D object detection.
        
\begin{figure*}[h]
	\centering
	\includegraphics[width=0.9\linewidth]{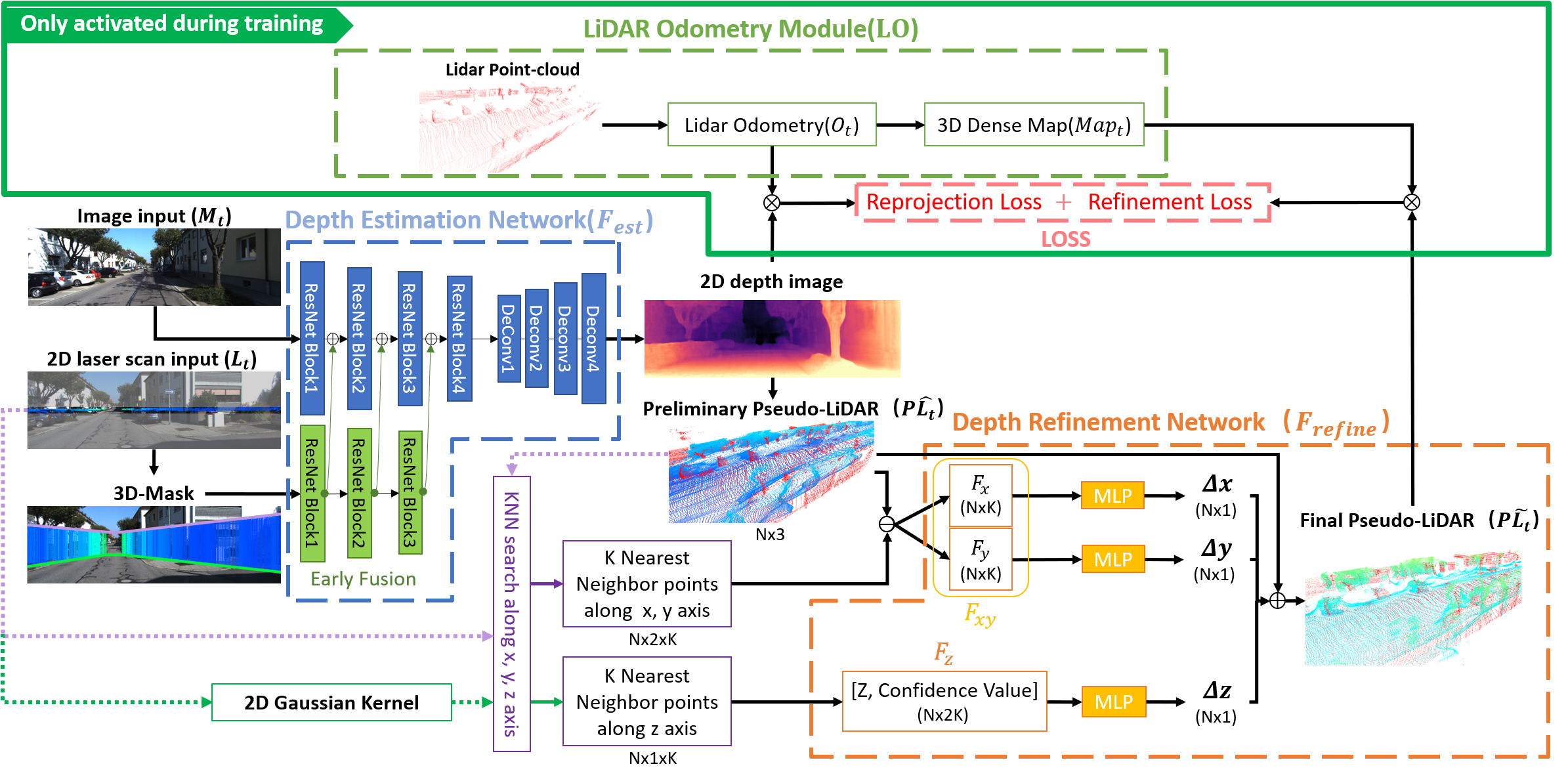}
	\caption{The framework of FusionMapping.
	The network consists of three core parts: \textit{LiDAR Odometry} provides relative pose transformation $O_t$ and a dense local map $\text{Map}_t$. This information is used to supervise the training process of the Depth Estimation Network $F_{est}$ and the Depth Refinement Network $F_{refine}$.
	The \textit{Depth Estimation Network} predicts preliminary pseudo-LiDAR $\hat{PL}_t$ from a monocular image $M_t$ and a 2D laser scan $L_t$.
	The \textit{Depth Refinement Network} further refines the pseudo-LiDAR based on $\hat{PL}_t$ and $L_t$. 
	Let $N$ represent the number of predicted points in $\hat{PL}_t$, and $K$ represent the number of neighbours for each predicted point. $F_x$, $F_y$ and $F_z$ are the extracted features for offset estimation.}
	\label{fig:framework}
\end{figure*}
        
    \subsection{Fusion-based Depth Prediction}
        Recently, researchers try to fuse different sensor information to improve the estimates of depth information.
        Ma et al. \cite{Sparse:mal2018sparse} use RGB images together with sparse depth information to train a bottleneck network architecture.
        Comparing to imagery only methods, their approach generates a better depth estimation results. Nevertheless, such 3D LiDAR devices are still expensive and prohibitive for most robotics platforms.
        Park et al. \cite{Sparse:park2018high} apply a calibration network and a depth fusion network to fuse uncalibrated LiDAR point-cloud information into stereo camera images, performing high precision depth estimation.
        Besides these sensor-based fusion approaches, other methods fuse post-generated information with RGB images. 
        For example, Yang et al. \cite{Sparse:yang2019bayesian} fuse an uncertainty map into a depth map for monocular video depth estimation.
        \\
    
    However, even with the assistance of LiDAR inputs, none of the above methods address our main concerns, \textit{long tails} and \textit{local misalignments}.
    Besides, no work has investigated fusing 2D laser scans and monocular images to generate better depth estimation.
    In the next section, we introduce our FusionMapping framework, which uses images and 2D laser scans to alleviate the above challenges.

\section{Method}

    At time stamp $t$, with a given monocular image $M_t$, 2D laser scan $L_t$, odometry $O_t$, and dense local map $\text{Map}_t$, FusionMapping predicts the preliminary pseudo-LiDAR $\hat{PL}_t$ and refined pseudo-LiDAR $\tilde{PL}_t$ with the following:
    \begin{equation}
        \hat{PL}_t = F_{est}(M_t, L_t),
        \label{eq:monodepth}
    \end{equation}
    \begin{equation}
        \tilde{PL}_t = F_{refine}(\hat{PL}_t, L_t),
        \label{eq:2d_constrains}
    \end{equation}
    where $F_{est}$ and $F_{refine}$ are the pseudo-LiDAR estimation network and refinement network respectively. 
    $O_t$ is the pose information generated by the LiDAR Odometry Module(LO). It is used during training of the depth estimation network($F_{est}$). 
    A dense local map $\text{Map}_t$ is used during training of the depth refinement network($F_{refine}$).
    As shown in Figure~\ref{fig:framework}, FusionMapping includes the following three parts:
    \begin{enumerate}
        \item LiDAR Odometry Module (LO): To provide accurate relative pose transformation for $F_{est}$ and 3D point-cloud reference for $F_{refine}$, we apply LiDAR odometry to perform pose estimation $O_t$, and to generate a dense LiDAR point-cloud $\text{Map}_t$. This module is only used during training time.
        \item Depth Estimation Network: The depth estimation network $F_{est}$ has a self-supervised network structure which takes $M_t$ and $L_t$ as inputs, and outputs a preliminary depth prediction $\hat{PL}_t$, which is trained under the guidance of LiDAR Odometry $O_t$.
        \item Depth Refinement Network ($F_{refine}$):
        Given the extracted pseudo-LiDAR $\hat{PL}_t$ and the 3D geometry reference $Map_t$, we extract the geometry constraints from the 2D laser scan $L_t$ to refine the 3D pseudo-LiDAR point-cloud.
    \end{enumerate}

    \begin{figure*}[h]
    	\centering
    	\includegraphics[width=\linewidth]{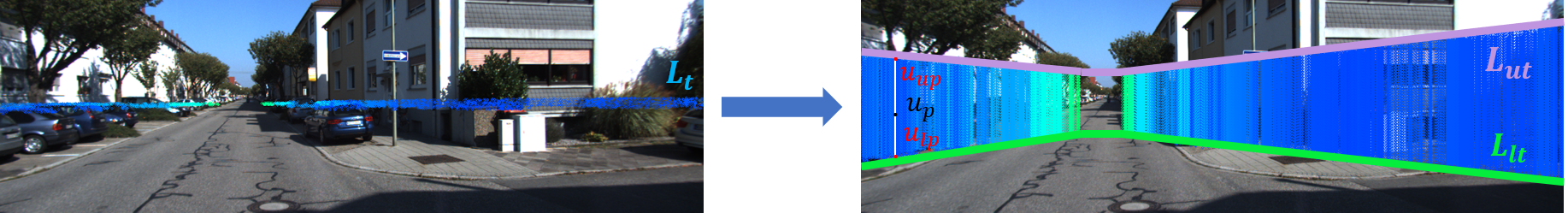}
    	\caption{The feature fusion operation in the Depth Estimation Network $F_{est}$.
    	The left image shows a 2D laser scan $L_t$ and frontal image $M_t$. 
    	In the right image, the upper purple line and lower green line are the upper boundary ${L_{u}}_t$ and lower boundary ${L_{l}}_t$ respectively, and the middle area is the 3D-masked area according to (\ref{eq:cond_depth}).}
    	\label{fig:early_fusion}
    \end{figure*}

    \subsection{LiDAR Odometry Module}
        \label{sec:odometry}
        The original \textit{Monodepth2}~\cite{PSEUDO:monodepth2} method utilizes per-pixel photometric loss between the input image and a neighbouring frame during training. Thus, it relies on an estimated relative pose transformation to predict the depth information, where the pose estimation is trained on the \textit{KITTI odometry} dataset. 
        However, for unknown datasets (such as in indoor environments), the accuracy of Monodepth2's pose estimation will decrease.
        To enable the training procedure to the unknown datasets, we apply a LiDAR Odometry module to generate reference odometry and a dense point-cloud.
        The LiDAR Odometry module could output accurate odometry information with the given 3D LiDAR point-cloud and IMU data.
        For more details on LiDAR odometry, we refer the reader to~\cite{LOAM:zhang2014loam}.
        
        Based on the relative pose transformation $O_{t}$，we generate a local dense point-cloud $\text{Map}_t$ by caching the five most recent 3D LiDAR point-cloud scenes, and cropping the 3D LiDAR point-clouds based on their relative distance to the mobile robot body center,
        \begin{align}
            \text{Map}_t=\{p\mid dist(p,c) \leq thresh, \nonumber \\
                  p \in [D_t, D_{t-1},...D_{t-4}]\},
            \label{eq:local_map}
        \end{align}
        where $p=[p_{x}, p_{y}, p_{z}]$ is a 3D point, $c$ is the robot center, $[D_t, D_{t-1},...D_{t-4}]$ are five consecutive point-clouds, and $thresh$ is a distance threshold.
        The cached local dense point-cloud $\text{Map}_t$ is used to supervise the learning of $F_{refine}$.
    
    \subsection{Depth Estimation Network}
        We adopt \textit{Monodepth2} as our depth estimation backbone.
        However, while \textit{Monodepth2} requires a relative pose prediction network to guide the depth prediction, we instead use the LiDAR odometry $O_t$ for this guidance.
        We apply a 3D-Mask operation to fuse the 2D laser scan in the first three convolution layers.
        
        \subsubsection{3D-Mask}
        As shown in Figure~\ref{fig:early_fusion} (left image), we take a 2D laser scan $L_t$ that is $1.2$m above the ground. 
        Typically, we observe that many objects of interest on the road are around $1.8\sim2.0$ m from the ground.
        To construct a 3D-Mask on the front view, we set a lower boundary ${L_{l}}_t$ (in Figure~\ref{fig:early_fusion}, the green line in the right image) and an upper boundary ${L_{u}}_t$ (the purple line in the right image) based on $L_t$
        \begin{equation}
            {L_{l}}_t=[p_l \mid p_{z} = p_{z}-1.2, p \in L_t],
            \label{eq:boundary_lower}
        \end{equation}
        \begin{equation}
            {L_{u}}_t=[p_u \mid p_{z} = p_z+0.8, p \in L_t].
            \label{eq:boundary_upper}
        \end{equation}
        where $p=[p_{x}, p_{y}, p_{z}]$ is a 3D point obtained from the 2D laser scan $L_t$.
        Based on the intrinsic and extrinsic matrix, the 3D points $\{p_{x}, p_{y}, p_{z}\mid p \in [{L_{l}}_t, {L_{u}}_t]\}$ in these boundaries are transformed into 2D image indexes $\{[u_p,v_p] \mid p \in [{L_{l}}_t, {L_{u}}_t]\}$ by
        \begin{equation}
          s
          \begin{bmatrix}
            u_p \\
            v_p \\
            1 \\
          \end{bmatrix} 
          = I_{intrinsic} \cdot I_{extrinsic} \cdot
          %
          \begin{bmatrix}
            p_x \\
            p_y \\
            p_z \\
            1 \\
          \end{bmatrix}
          \label{eq:2d_to_3d}
        \end{equation}
        where both the intrinsic matrix $I_{intrinsic}$ and extrinsic matrix $I_{extrinsic}$ are calibrated offline, and $s$ is the depth of $u_p,v_p$ in the camera frame.
        Given a front view image of size $[M, N, 3]$, we construct a 3D-Mask with size $[M, N, 1]$,
        \begin{equation}
            3DMask(u_p,v_p)=
            \begin{cases}
              s, & \text{if}\ u_{lp} \leq u_p \leq u_{up} \\
              0,       & \text{otherwise}
            \end{cases},
           \label{eq:cond_depth}
        \end{equation}
        where $u_{lp}, u_{up}$ are the corresponding lower and upper boundary points of $(u_p,v_p)$ on $L_{lt}$ and $L_{ut}$.
        An example of this mask is shown in Figure~\ref{fig:early_fusion}.
        We apply multiple early fusion layers between the  camera image and the additional 3D-Mask channel as shown in the $F_{est}$ network in Figure~\ref{fig:framework}. 
        Specifically, we pass the 3D-Mask into a ResNet~\cite{RES:he2016deep} that has the same network structure as the image input branch. Then, we sum the intermediate output of the ResNet block with those of the image input branch.
        In Section \ref{sec:Experiment}, we present experiments on different fusion configurations. 
        The loss formulation for the prediction of $\hat{PL}_t$ is that of the original \textit{Monodepth2}~\cite{PSEUDO:monodepth2}.

    \subsection{Depth Refinement Network}
        Although $\hat{PL}_t$ is represented as a 2D depth image, it is important to evaluate depth information in 3D. As we described before, when we project 2D depth images to 3D, they often have \textit{long tails} and \textit{misalignment} problems. Thus, for the first step of our depth refinement network, we project $\hat{PL}_t$ onto a 3D space. Next we describe how to utilize 2D laser scan for further refinement.
        Note that the 2D laser scan cannot directly provide a 3D reference for the pseudo-LiDAR $\hat{PL}_t$, so additional preprocessing of the 2D laser scan is needed.
        
        \begin{figure}[h]
        	\centering
        	\includegraphics[width=\linewidth]{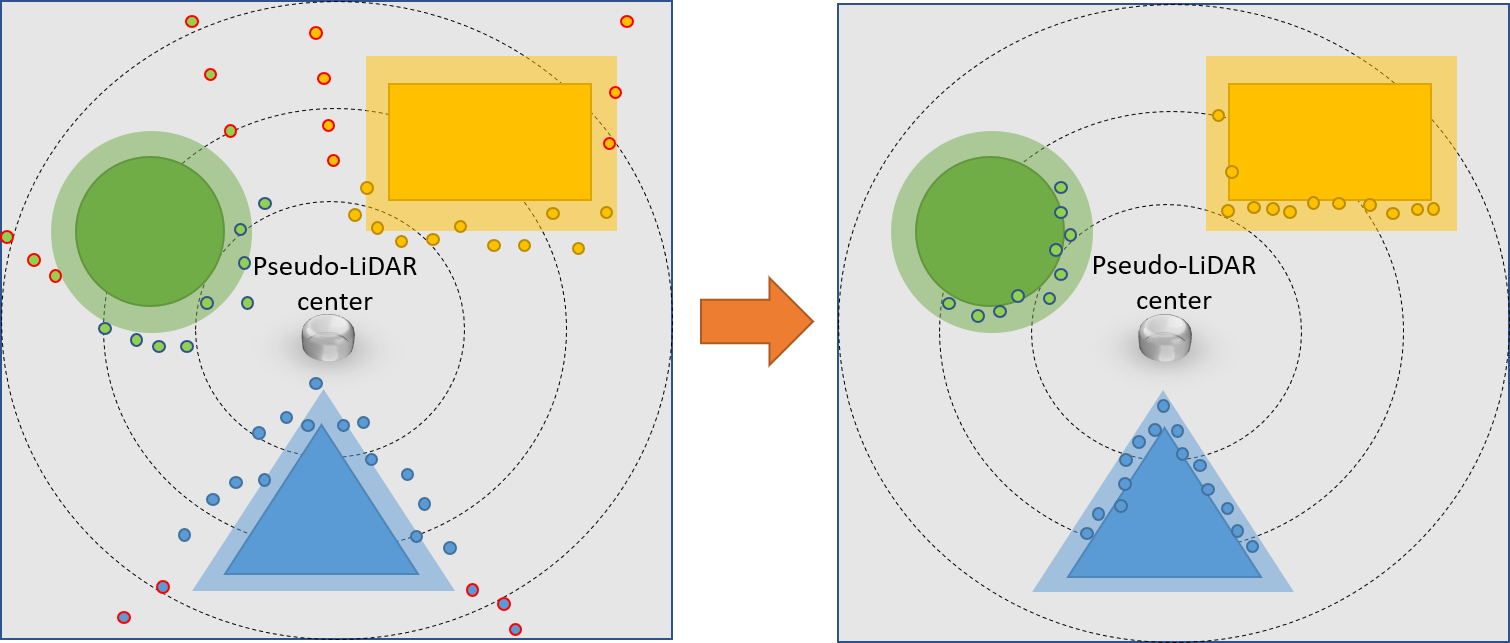}
        	\caption{The refinement operation.
            The preliminary pseudo-LiDAR $\hat{PL}_t$ (left) may have \textit{long tails} (red circles) and \textit{misalignments}.
            After refinement (right), the \textit{long tails} have been removed by $F_z$ in Eq.~\ref{eq:rf_z}, and the \textit{misaligned} points are regressed to their neighboring objects with Eq.~\ref{eq:rf_xy}.}
        	\label{fig:refinement}
        \end{figure}

        As depicted in Figure~\ref{fig:refinement}, we apply a Gaussian kernel on the 2D laser scan $L_t$ (represented by solid objects in the figure)) to represent a confidence map (represented by translucent objects in the figure), which is used to predict whether each point in $\hat{PL}_t$ belongs to an object boundary.

        \begin{table*}[t]
        \small
        \caption{Ablation analysis on the \textit{KITTI} dataset with various fusion configurations.}
        \begin{center}
        \begin{tabular}{|c||*{6}{c|}|c|c|}
        	\hline
        	& \multicolumn{6}{c||}{2D metric} & \multicolumn{2}{c|}{3D metric} \\
            & \cellcolor{red!25}Abs Rel & \cellcolor{red!25}Sq Rel & \cellcolor{red!25}RMSE & \cellcolor{blue!25}$\sigma < 1.25$ & \cellcolor{blue!25}$\sigma < 1.25^2$ & \cellcolor{blue!25}$\sigma < 1.25^3$ & \cellcolor{green!25}EMD & \cellcolor{green!25}EFS\\
        	\hline
        	Monodepth2 & 0.141 & 0.720 & 4.248 & 0.810 & 0.961 & 0.946 & 25.302 & 0.389\\
        	\hline
        	FM: Early Fusion 1 & 0.118 & 0.576 & 4.733 & 0.835 & 0.966 & 0.943 & 22.258 & 0.386\\
        	\hline
        	FM: Early Fusion 2 & \textbf{0.104} & 0.515 & \textbf{3.541} & \textbf{0.892} & 0.979 & \textbf{0.955} & 19.094 & 0.375\\
        	\hline
        	FM: Early Fusion 3  & 0.106 & \textbf{0.510} & 3.634 & 0.886 & \textbf{0.980} & 0.949 & 21.626 & 0.383\\
        	\hline
        	FM: Early Fusion 2 + Refinement & NA & NA & NA & NA & NA & NA & \textbf{15.083} & \textbf{0.001}\\
        	\hline
        \end{tabular}
        \label{table:Ablation}
        \end{center}
        \end{table*}   

        \subsubsection{Misalignment Refinement}
        To conduct misalignment refinement, we construct 2D geometry constraints for the pseudo-LiDAR $\hat{PL}_t$.
        For each predicted 3D point $\hat{p}\in \hat{PL}_t$, with 2D K-nearest neighbor  searching on 2D raw laser scan, we can extract its 2D constraints feature $F_{xy}$ by
        \begin{align} F_{xy}&=[n_{xyi}.x-\hat{p}.x, n_{xyi}.y-\hat{p}.y \mid i=1...k],
            \label{eq:local_map}
        \end{align}
        where $n_{xyi}$ are the KNN neighbors on 2D laser scan points, $k$ is the number of neighbors. In practice we choose $k=9$.
        
        \subsubsection{Long Tails Refinement}
        To remove long tails, for each point $\hat{p}\in \hat{PL}_t$, we add a constrain on its height relative to neighboring points in $\hat{PL}_t$.
        We first retrieve $k$ nearest $\hat{p}\in \hat{PL}_t$ in the preliminary pseudo-LiDAR point set $\hat{PL}_t$. Then, we project these neighbors onto the confidence map generated from the 2D laser scan. Finally we estimate the height constraint features by
        \begin{align}
            F_{z} &=[n_{zi}.z-\hat{p}.z, bev_i \mid i=1...k],
            \label{eq:local_map}
        \end{align}
        where $n_{zi}$ are the KNN neighbors of $\hat{p}$ in the pseudo-LiDAR $\hat{PL}_t$, $k$ represents the number of neighbors, and $bev_i$ represents the confidence score of neighboring point $n_{zi}$.\hfill \break
    
    Based on the above feature extractions constraints on 2D geometry and height, the prediction for each point $\hat{p}$ is estimated by
    \begin{align}
        &[\tilde{p}.x, \tilde{p}.y] = [\hat{p}.x, \hat{p}.y] + \text{MLP}(F_{xy}) \label{eq:rf_xy} \\
        &\tilde{p}.z = \hat{p}.z + \text{MLP}(F_{z}). \label{eq:rf_z}
    \end{align}
    where \textit{MLP} is the output of a multi-layer perceptron. 
    With the 3D reference point-cloud retrieved from $\text{Map}_t$, the refinement loss $Loss_{refine}$ is given by
    \begin{align}
        &Loss_{refine}=\sum \norm{(\tilde{PL}_t(i)-W_{ij}\cdot \text{Map}_t(j)}^2 \nonumber \\
        &\text{s.t.}\; W_{ij}=
        \begin{cases}
          1,       & \text{if}\; \underset{j}{\arg} \min \norm{\tilde{PL}_t(i)-\text{Map}_t(j)}\\
          0,       & \text{otherwise}
        \end{cases}.
        \label{eq:local_map}
    \end{align}
    where $W_{ij}$ is the weighting between point $\tilde{PL}_t(i)$ and $Map_t(j)$, and also satisfies $\underset{j}{\sum} W_{ij}=1$.

\section{Experiments}
\label{sec:Experiment}
\begin{figure}[!t]
        	\centering
        	\includegraphics[width=1.0\linewidth]{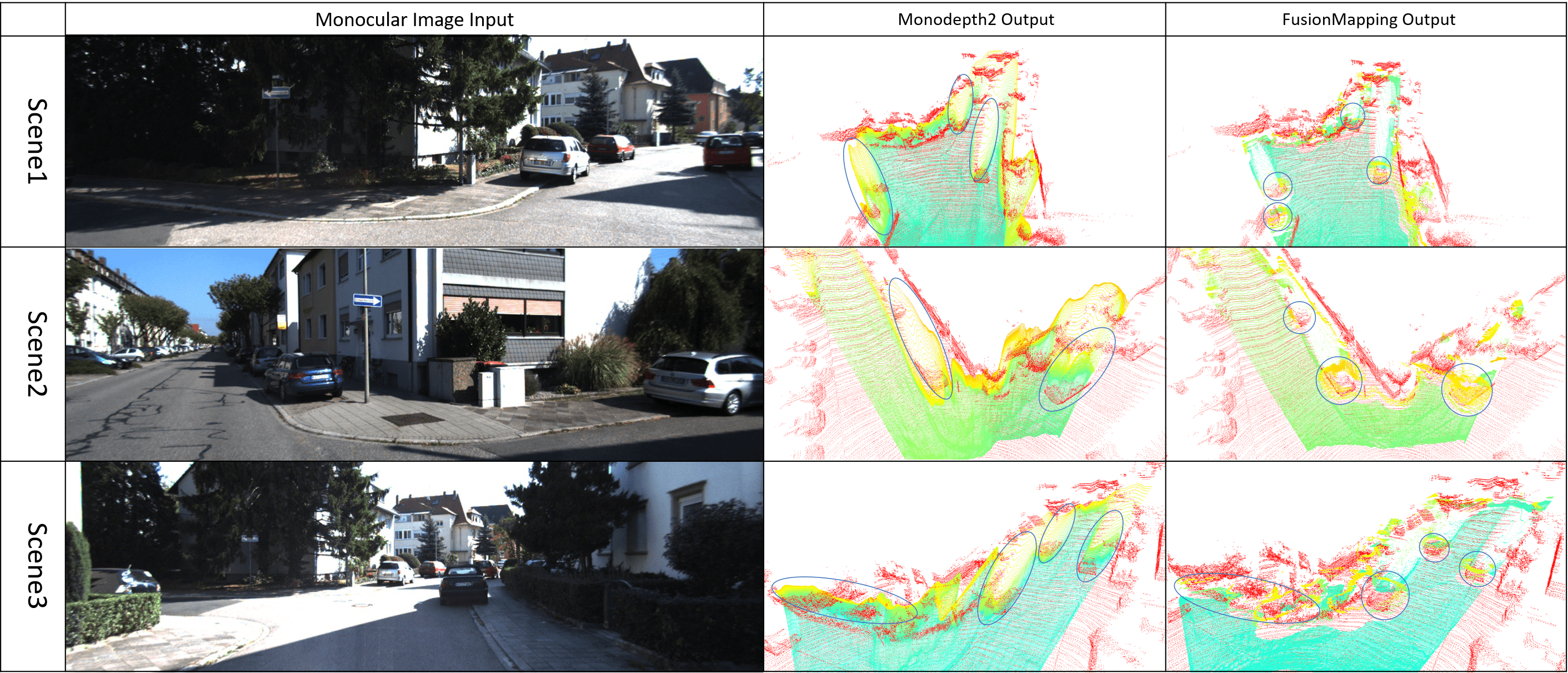}\\
        	\vspace{0.2cm}
        	\includegraphics[width=1.0\linewidth]{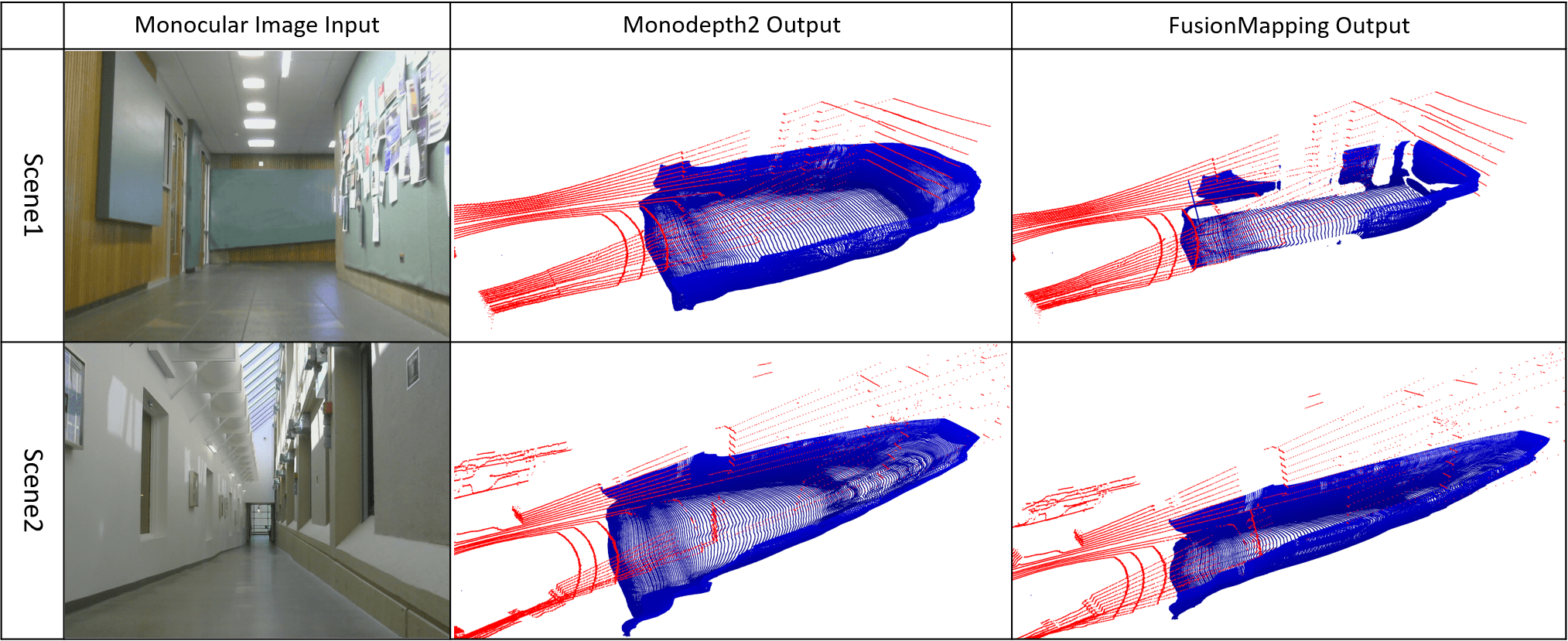}
        	\caption{Qualitative Analysis on the \textit{KITTI} and \textit{Mobile} robot datasets. 
        	The circled areas demonstrate the long tails and misaligned points.
        	In the \textit{KITTI} dataset, both the above problems could be solved by FusionMapping. 
        	In the \textit{mobile robot} dataset, the environment is comparatively clean and simple, and we see that the main problem with point-clouds output by Monodepth2 is the misalignment of walls. Our network can successfully decrease this misalignment.}
        	\label{fig:qual_kitti}
        \end{figure}

        To test the performance of our method, we trained and evaluated the network on two datasets. the first is the \textit{KITTI} odometry dataset, and the second is a dataset generated from a physical mobile robot.
        The details of these two datasets are as follows:
        \begin{itemize}
            \item The \textit{KITTI odometry dataset} is an Autonomous Vehicle dataset containing Velodyne-64 LiDAR point clouds, monocular images, stereo images, and relative pose information, all of which are generated in an urban environment. 
            Given the \textit{KITTI} odometry dataset, we generated 31,983 samples with the LiDAR odometry network. Of these, 21,951 samples were used in training, and the remaining for evaluation.
            \item Our \textit{mobile robot} dataset is an indoor environment dataset generated by our four-wheeled robot, which has one VLP-16 LiDAR device and one monocular camera. 
            We generate 15,291 samples with our LiDAR odometry module. Of these, 10,120 samples were used for training, and 5,000 for evaluation.
        \end{itemize}
        In the datasets, each sample contains the current image view $M_t$, relative pose between two images $O_t$, and dense local 3D point cloud $Map_t$.
        We compare our method with the  \textit{Monodepth2}. 
        In the following section, we conduct an ablation study on the \textit{KITTI odometry} dataset to evaluate the quantitative performance of different methods.
        In Figure~\ref{fig:qual_kitti}, we show the qualitative analysis on the \textit{KITTI} and \textit{mobile robot} dataset separately.
        This includes a study on the effects of different fusion configurations, and how well our method solves the \textit{long tails} and \textit{misalignment} problems.
        All experiments were conducted on an Ubuntu 18.04 system, with a single Nvidia 1080Ti GPU and 64 Gb RAM.
        
        
        \begin{figure}[t]
        	\centering
        	\includegraphics[width=\linewidth]{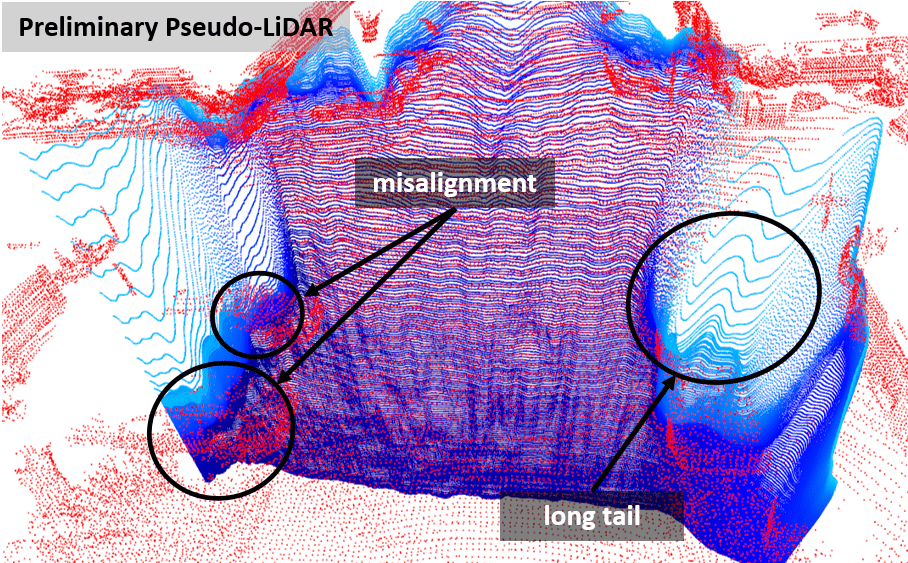}
        	\caption{``Long tails'' and ``misalignment'' problems are common in the original pseudo-LiDAR generation ($\hat{PL}_t$). The red point-cloud is the ground truth, and the blue point-cloud is the predicted preliminary pseudo-LiDAR $\hat{PL}_t$ }
        	\label{fig:long_tail}
        \end{figure} 
        
    \subsection{Ablation Study on the \textit{KITTI Odometry} Dataset}

        To investigate the performance of our early fusion layers and depth refinement network, we performed an ablation study, the results of which are shown in  Table~\ref{table:Ablation}.
        For quantitative analysis, we measured the prediction performance with the 2D depth metric from~\cite{METRIC:eigen2014depth}.
        We also introduced two additional 3D metrics to measure the quality of the generated pseudo-LiDAR: Earth Mover's Distance (EMD) and Euclidean Fitness Score (EFS)~\cite{PCL:rusu20113d}.
        To calculate the EMD loss, also known as the Wasserstein distance, we adopt the equations and implementation of \cite{feydy2018interpolating}. 
        The Euclidean Fitness Score (EFS) is the sum of the squared distance of $\tilde{PL}_t$ and $Map_t$ provided in the Point cloud library~\cite{PCL:rusu20113d}. 
        Our baseline is the original \textit{Monodepth2} method. 
        Compared to the baseline, our fusion-based method results in higher quality depth prediction with respect to above mentioned 2D depth metrics.
        Further, our method significantly improved the quality of the pseudo-LiDAR with respect to the 3D metrics. 
        Next we will analyze the performance of different fusion configurations in the Depth estimation and refinement networks separately.
        
        \begin{figure}[t]
        	\centering
        	\includegraphics[width=\linewidth]{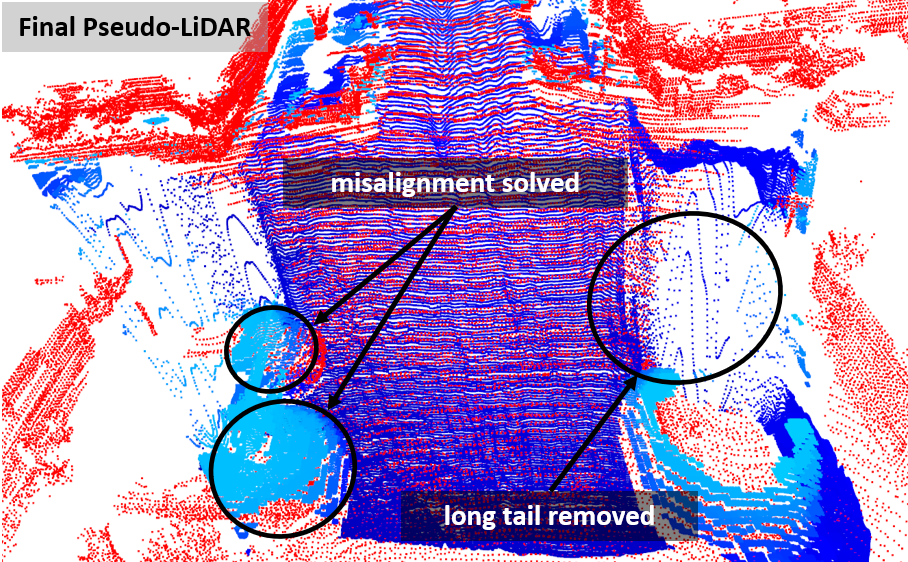}
        	\caption{FusionMapping is able to alleviate the problems common in pseudo-LiDAR ($\tilde{PL}_t$) using the Depth Refinement Network ($F_{refine}$). After refinement, most long tails are removed and misaligned points are aligned to object boundaries.}
        	\label{fig:miss_matching}
        \end{figure} 
    
        \subsubsection{Depth Estimation Network}
        In order to find out the effects of early fusion layers, we performed tests on three different levels of fusion configurations: from shallow fusion to deep fusion. Early fusion 1 only fuses the output of the first block of 3D-Mask branch and input image branch, early fusion 2 fuses the outputs of the first two blocks, and early fusion 3 fuses the outputs of the first three blocks. 
        Compared to the our baseline, early fusion layers improve depth estimation accuracy.
        We found that deeper fusion layers (Early fusion 2, 3) resulted in higher accuracy than the shallow layer configurations, as shown in Table~\ref{table:Ablation}, we compared the original \textit{Monodepth2} method with FusionMapping (FM) method of different configurations.
        The performance measurements are evaluated under the 2D depth metrics~\cite{METRIC:eigen2014depth}, Earth Mover's Distance (EMD),and Euclidean Fitness Score (EFS) 3D metrics.
        
        \subsubsection{Depth Refinement Network}
        To evaluate the effects of adding a depth refinement network on top of the depth estimation network, we introduce two metrics: EMD and EFS. These two metrics evaluate a matching loss between the predicted point-cloud $\hat{PL}_t$ and $Map_t$. 
        Since $F_{refine}$ transforms a 2D depth map into a 3D point-cloud, we do not evaluate our final output on 2D metrics. 
        Early Fusion 2 and Early Fusion 3 have comparable performance under 2D metrics. To evaluate our final network, we only choose Early Fusion 2 with $F_{refine}$ (FusionMapping).

        We found that the FusionMapping significantly improved the 3D mapping performance over the variations.
        In particular, all methods without the refinement network performed worse on the 3D metric. This shows that the 2D constraint on the $XY$ plane and the height constraint can further refine the original preliminary pseudo-LiDAR.
    
    \subsection{\textit{Long Tails} Analysis}
        Figure~\ref{fig:long_tail} shows the \textit{long tails} problem when converting predicted 2D depth image $\hat{P}_t$ to 3D point-cloud.
        Most objects have noisy depth predictions on the boundaries, which are mainly caused by the blurring effect of multi-layer convolution operations.
        Given height constraints $F_{z}$, the height refinement operation in (\ref{eq:rf_z}) effectively reduces noisy tail points, while the geometry structure of the original objects, such as car and traffic light pole (see Figure~\ref{fig:miss_matching}), are preserved.

    \subsection{\textit{Misalignment} Analysis}
        We can see in Figure~\ref{fig:long_tail} that points on the vehicles' front boundaries and side walls usually are misaligned compared to the real 3D geometry.
        Given 2D plane constraints $F_{xy}$, the 2D refinement operation in {\ref{eq:rf_xy}} eliminates such offsets.
        There are also fewer misaligned points on distant objects.

    \subsection{Qualitative Analysis}
        For both \textit{KITTI} and \textit{Mobile robot} dataset, we qualitatively analyzed the point cloud outputs, for example, those shown in  Figure~\ref{fig:qual_kitti}.
        Due to the 2D constraints on the $XY$ plane, noisy points around vehicles and walls are pulled onto the objects' boundaries. This is especially important for depth estimation in the indoor environment, where there are many narrow corridors with few features. Nevertheless, the 2D laser scan is more reliable indoors than outdoors. Thus our method has advantages for indoor mapping as well.
        
    \subsection{Limitations of Current Work}
    \label{subsec:limitation}
        From the qualitative and quantitative analysis in the above experiments, the proposed FusionMapping method demonstrated its ability in pseudo-LiDAR refinement.
        However, we noticed a few limitations to our method. 
        The network, when trained on an indoor environment, does not generalize to outdoor data well, and vice versa. As such, the network must be trained for each new environment. Finally, our method does not account for  changes in lighting condition and dynamic objects, which will introduce noise to depth predictions, especially in indoor environments.
    
\section{Conclusion}
    In this paper, we introduced a novel depth prediction framework, FusionMapping, which could predict accurate depth information with the assistance of monocular images and 2D laser scans.
    This framework contains two major modules: 
    1) a depth prediction network with multi-sensor fusion for preliminary point-clouds estimation; 
    2) and a depth refinement network with geometry constraints for point-clouds refinement.
    In the training procedure, we utilized a LiDAR odometry module to generate relative pose estimation and local dense maps to supervise the training process.
    Finally, we investigated the performance of our method on the \textit{KITTI odometry} dataset and a \textit{mobile robot} platform. 
    We found our method improved the depth prediction over original image-based approach significantly.
    
    In future work, we will focus on how to improve the generalization ability of our method for varying lighting conditions, dynamic environments and unseen environments. 

\bibliographystyle{plain}
\bibliography{ref}
\raggedbottom
\end{document}